\pdfoutput=1

\documentclass[11pt]{article}

\usepackage{siunitx}
\usepackage[]{acl2022}

\usepackage{times}
\usepackage{latexsym}

\usepackage{amsmath}
\usepackage{mathtools}
\usepackage{tikz}
\usepackage{mathdots}
\usepackage{yhmath}
\usepackage{adjustbox}
\usepackage{multirow}
\usepackage{graphicx}
\usepackage{booktabs}
\usepackage{nccmath}
\usepackage{bbm}

\usepackage{physics}
\usepackage{amsmath}
\usepackage{tikz}
\usepackage{mathdots}
\usepackage{yhmath}
\usepackage{cancel}
\usepackage{color}
\usepackage{array}
\usepackage{multirow}
\usepackage{amssymb}
\usepackage{tabularx}
\usepackage{extarrows}
\usepackage{booktabs}
\usepackage{algorithm}
\usepackage{algpseudocode}
\usetikzlibrary{fadings}
\usetikzlibrary{patterns}
\usetikzlibrary{shadows.blur}
\usetikzlibrary{shapes}
\DeclareMathOperator*{\argmax}{argmax} 
\DeclareMathOperator*{\argmin}{argmin} 
\makeatletter
\newcommand{\multiline}[1]{%
  \begin{tabularx}{\dimexpr\linewidth-\ALG@thistlm}[t]{@{}X@{}}
    #1
  \end{tabularx}
}
\makeatother

\algblock{Input}{EndInput}
\algnotext{EndInput}
\algblock{Output}{EndOutput}
\algnotext{EndOutput}

\usepackage[T1]{fontenc}

\usepackage[utf8]{inputenc}

\usepackage{microtype}

\title{On the Calibration of Pre-trained Language Models using Mixup \\Guided by Area Under the Margin and Saliency}

\author{Seo Yeon Park and Cornelia Caragea\\
Computer Science  \\
  University of Illinois Chicago \\
  {\tt spark313@uic.edu} \mbox{     } \mbox{     } \mbox{     }
    {\tt cornelia@uic.edu}
}

\begin{document}
\maketitle
\begin{abstract}
A well-calibrated neural model produces confidence (probability outputs) closely approximated by the expected accuracy. While prior studies have shown that mixup training as a data augmentation technique can improve model calibration on image classification tasks, little is known about using mixup for model calibration on natural language understanding (NLU) tasks. In this paper, we explore mixup for model calibration on several NLU tasks and propose a novel mixup strategy for pre-trained language models that improves model calibration further. Our proposed mixup is guided by both the Area Under the Margin (AUM) statistic \cite{NEURIPS2020_c6102b37} and the saliency map of each sample \cite{simonyan2013deep}. Moreover, we combine our mixup strategy with model miscalibration correction techniques (i.e., label smoothing and temperature scaling) and provide detailed analyses of their impact on our proposed mixup. We focus on systematically designing experiments on three NLU tasks: natural language inference, paraphrase detection, and commonsense reasoning. Our method achieves the lowest expected calibration error compared to strong baselines on both in-domain and out-of-domain test samples while maintaining competitive accuracy.
\end{abstract}

\section{Introduction}
Training a well-calibrated classifier that produces a match between confidence (the probability output that a model assigns to a prediction) and correctness (accuracy), is important in modern neural networks. As an example, if an AI-based application {\em knows what it does not know}, or in other words, the chance that the current prediction is wrong, a human is more helpful to correct the error.
However, many works reveal that current deep neural networks are prone to over-confidence, which implies that the models' confidence is not reliable \cite{guo2017calibration}. This is a critical issue on the deployment of AI-based user applications such as the healthcare domain \cite{zhu2018clinical,info:doi/10.2196/14830} or safety-critical domain \cite{sarabadani-2019-detection} due to the problem of prediction trustworthiness. 

Recently, the study of calibration on neural network models especially on natural language processing tasks has started to receive attention. 
To overcome the problem of miscalibration, numerous suggestions on how to address it have been proposed.
For example, \citet{guo2017calibration} revealed that using temperature scaling before the final softmax layer reduces calibration errors.
\citet{muller2019does}, \citet{kumar2019calibration}, and \citet{wang-etal-2020-inference} found that label smoothing and its variants yield better calibration for neural machine translation. 
\citet{desai2020calibration} also reported that the aforementioned miscalibration correction methods can be applied to calibrate pre-trained language models which are often miscalibrated potentially due to over-parameterization. 

Mixup \cite{zhang2017mixup} is a data augmentation method for deep neural networks in which additional samples are generated during training by combining random pairs of training inputs and their associated labels. While simple to implement, mixup has been shown to improve both predictive performance and model calibration, particularly on image classification tasks due to its regularization effect through data augmentation \cite{singh2019mixup}. 
The recent success of mixup on image classification has led to the development of various mixup strategies for NLU especially those that use hidden state representations \cite{guo2019augmenting,chen-etal-2020-mixtext,zhang-etal-2020-seqmix,sun-etal-2020-mixup, kong-etal-2020-calibrated,yin-etal-2021-batchmixup}. 
However, most prior works on NLU focus on performance improvement using mixup rather than model calibration. Despite its benefits for calibration, a mixup for correcting miscalibrated predictions is still an under-explored topic in NLU.
While \citet{kong-etal-2020-calibrated} explored BERT \cite{devlin2019bert} calibration using mixup for both in-domain and out-of-domain, they only focused on generating mixup samples by utilizing the distance between instances in the feature space.
In contrast, we propose a novel mixup method, in which we first leverage the behavior of a model on individual samples during training (training dynamics), which can reveal samples with distinct pronounced characteristics---whether they are easy-to-learn or hard-to-learn/ambiguous for the model, and then we generate mixup samples by mixing easy-to-learn with hard-to-learn/ambiguous samples according to their similarity/dissimilarity provided by saliency maps. Saliency maps capture how much each data portion contributes to the final classification decision of a sample \cite{simonyan2013deep}. Intuitively, easy-to-learn samples help with model optimization, whereas hard-to-learn or potentially ambiguous samples are essential for learning since they are the most challenging for the model \cite{swayamdipta2020dataset}, and mixing them using saliency maps can yield better calibrated models (more realistic model confidence), e.g., mixing easy-to-learn with hard-to-learn/ambiguous samples by similarity in saliency maps can benefit in-domain calibration and by dissimilarity can benefit out-of-domain calibration. 
To monitor training dynamics, we use the Area Under the Margin (AUM) statistic \cite{NEURIPS2020_c6102b37} which measures how different a true label for a sample is compared to a model's \textit{beliefs} at each epoch and is calculated as the average difference between the logit values for a sample’s assigned class and its highest non-assigned class across training epochs.

Moreover, we combine our mixup with well-known miscalibration correction methods such as label smoothing and temperature scaling \cite{guo2017calibration} to investigate their impact on our proposed mixup.
We conduct a comprehensive set of experiments using BERT \cite{devlin2019bert} and RoBERTa \cite{liu2019roberta} to show the efficacy of our mixup approach by testing on three NLU tasks: natural language inference, paraphrase detection, and commonsense reasoning. We achieve the lowest Expected Calibration Error (ECE) without accuracy drops in comparison with strong baseline methods.  
Our contributions are as follows:

\begin{itemize}
    \item We propose a novel mixup method which is guided by AUM and saliency signals and is targeted at improving model calibration. 
    Specifically, we compare logits to categorize samples into two sets (i.e., a set of easy-to-learn samples and another set of hard-to-learn/ambiguous samples), and interpolate samples across these two sets by finding the most similar and most dissimilar samples from the other set while leveraging saliency (to compute sample similarities) for pre-trained language models' calibration on in-domain and out-of-domain data. 
    
    \item We combine our method with miscalibration correction techniques (i.e., label smoothing, temperature scaling) to investigate their impact on our proposed mixup.
    
    \item  We conduct comprehensive experiments showing that our method achieves the lowest expected calibration errors (ECEs) on both in-domain and out-of-domain samples compared with strong baselines without accuracy drops on multiple NLU tasks, namely, natural language inferences, paraphrase detection, and commonsense reasoning.  
\end{itemize}

\section{Related Work}
\paragraph{Model Calibration} Calibration on NLU tasks has been widely studied in related literature. 
\citet{nguyen-oconnor-2015-posterior} provided the method of how to analyze the calibration of non-neural NLP models.
\citet{guo2017calibration} examined the calibration of modern deep neural networks and revealed that techniques such as temperature scaling and dropout affect the calibration on binary/multi-class classification tasks. 
\citet{wang2020inference} investigated the calibration of neural machine translation models and found that inference suffers from serious miscalibration. 
\citet{jagannatha-yu-2020-calibrating} demonstrated that neural networks show high calibration error on structured predictions such as NER, POS, and QA, and proposed to use a binary class forecaster to calibrate the predictor confidence for a defined output entity of interest. 
\citet{desai2020calibration} explored pre-trained language models' calibration in combination with temperature scaling and label smoothing both on in-domain and out-of-domain datasets.
\citet{jung-etal-2020-posterior} jointly optimized two objectives (a cross-entropy loss and a calibration loss) and directly penalized the difference between the predicted and the true posterior probabilities dynamically over the training steps.
\citet{he-etal-2021-joint} obtained better calibration on natural language understanding tasks by augmenting and training the classifier jointly with an energy-based model using noise-contrastive estimation. 

\paragraph{Mixup} 
Mixup \cite{zhang2017mixup} is a method for data augmentation in which additional samples are generated during training by convexly combining random pairs and their associated labels, and aims to alleviate overfitting. \citet{verma2019manifold} showed that manipulating hidden representations rather than manipulating input-level features on mixup results in better regularization effects due to the fact that it encourages the neural network to focus more on representations of the real training examples in a low dimensional subspace.
Many works have empirically noticed regularization effects that improve model performance on deep neural networks. 
For example, \citet{guo2019augmenting} explored the NLU specific mixup strategy by using sentence and word embeddings on CNNs and LSTMs to add performance gains in supervised text classification. 
\citet{chen-etal-2020-mixtext} proposed mixup for semi-supervised learning in which labeled and unlabeled samples are interpolated with their hidden representations to improve the performance of text classification.
\citet{zhang-etal-2020-seqmix} explored mixup for sequence labeling tasks with active learning to improve the performance of supervised sequence labeling tasks. 
\citet{yin-etal-2021-batchmixup} proposed mixup that interpolates every instance in a mini-batch to boost the performance of NLU tasks on the pre-trained language model RoBERTa \cite{liu2019roberta}. 
Similar to us, \citet{yoon-etal-2021-ssmix} explored mixup by incorporating saliency signals to generate augmented samples. 
Precisely, they use saliency signals to select a span of text from one sample to be replaced with another text span from another sample.
However, in contrast, our method first divides data samples into two categories (easy-to-learn and hard-to-learn/ambiguous categories) according to their AUM \cite{NEURIPS2020_c6102b37} distribution monitored over training epochs and then uses saliency to find the most similar/dissimilar samples across these two data categories.

\vspace{-1mm}
Recently, several works started to explore mixup for NLU model calibration. 
For example, \citet{singh2019mixup} investigated the impact of mixup for model calibration of NLU but only explored in-domain settings with simple deep learning architecture such as CNNs. 
\citet{kong-etal-2020-calibrated} explored BERT calibration using mixup as a regularization component on in-domain and out-of-domain. However, their mixup method only relied on the feature space distance between samples.
In contrast, we explore a novel mixup method in which we categorize the training samples into two sets using AUM \cite{NEURIPS2020_c6102b37} and combine samples across these two sets based on saliency signals, for in-domain and out-of-domain model calibration. 

\section{Approach}
\subsection{Mixup}
\paragraph{Background} Let $\mathcal{D}_{train}=\{({x}_i,y_i)\}_{i=1,\cdots,n}$ be a training set and $f$ a language model. 
Mixup training generates vicinity training samples according to the rule introduced in \citet{zhang2017mixup}:  
\begin{equation}
\begin{aligned}
    \tilde{x} = \lambda x_{i} + (1-\lambda)x_{j} \\
    \tilde{y} = \lambda y_{i} + (1-\lambda)y_{j}
\end{aligned}
\label{eq:1}
\end{equation}
where $x_{i}$ and $x_{j}$ are two randomly sampled input points, $y_{i}$ and $y_{j}$ are their associated one-hot encoded labels, and $\lambda$ is a mixing ratio sampled from a Beta($\alpha$, $\alpha$) distribution with a hyper-parameter $\alpha$.
In mixup, training data is augmented by linearly interpolating training samples in the input space.

\subsection{Proposed Approach} 
We propose a mixup method targeted at improving model calibration that synthesizes samples guided by the Area Under the Margin (AUM) \cite{NEURIPS2020_c6102b37} and saliency \cite{simonyan2013deep}. 

\paragraph{Data Categorization} In our method, we first categorize $\mathcal{D}_{train}$ into two sets (a set of easy-to-learn samples and a set of hard-to-learn/ambiguous samples) according to the AUM of each sample.
Given a sample $({x}_i,y_i)$, we compute $AUM({x}_i,y_i)$ as the area under the margin averaged across all training epochs $T$. Specifically, at some epoch $t\in T$, the margin is defined as:
\begin{equation}
    M^{t}(x_i, y_i) =  z_{y_i} - max_{y_i!=k}(z_{k})
\end{equation}
\noindent
where $M^{t}(x_i, y_i)$ is the margin of example $x_i$ with gold label $y_i$, $z_{y_i}$ is the logit corresponding to the gold label $y_i$, and $max_{y_i!=k}(z_{k})$ is the largest {\em other} logit corresponding to label $k$ not equal to $y_i$. Precisely, the margin measures how different a gold label is compared to a model's \textit{beliefs} at each epoch $t$. The AUM of $(x_i,y_i)$ across all epochs is:

\begin{equation} 
    AUM(x_i, y_i) =  \frac{1}{T} \sum_{t=1}^{T}M^{t}(x_i, y_i)
    \label{eq:aum}
\end{equation}

Intuitively, the samples with high AUM are easy-to-learn (the model's belief matches the gold label), but they are essential for model optimization, while the samples with low AUM are hard-to-learn or ambiguous (and hence they are the most challenging for the model), but they are essential for learning.
Our proposed mixup method first splits $\mathcal{D}_{train}$ into two data categories depending on whether the AUM value is high or low, namely, $\mathcal{D}_{high}$ and $\mathcal{D}_{low}$. In experiments, we compute the median AUM over the entire training samples and use it as a threshold to split the dataset. If a sample has a lower AUM than the threshold, we add the sample to $\mathcal{D}_{low}$, otherwise we add it to $\mathcal{D}_{high}$. Accordingly, we balance $\mathcal{D}_{high}$ and $\mathcal{D}_{low}$, but other splits are possible.
We then conduct a mixup operation by referring to each other set.
Mixing easy-to-learn and hard-to-learn adjusts the difficulty of samples and hence adjusts models' confidence according to samples' difficulties and yields better calibrated models. 
The data categorization step is summarized in Algorithm \ref{alg:aum}. 

\begin{algorithm}[t]
\flushleft
\caption{: Identify high/low AUM samples }\label{alg:aum}
\begin{algorithmic}[1]
    \Require  \begin{varwidth}[t]{\linewidth}
               $\mathcal{D}_{train}=\{({x}_i,y_i)\}_{i=1,\cdots,n}$; model $f$ \par
              \end{varwidth}
    \Function{Data-Categorization}{$\mathcal{D}_{train}$}
        \State $\mathcal{D}_{high} \gets \emptyset, \mathcal{D}_{low} \gets \emptyset$
        \State \multiline{Train $f$ for $T$ epochs and compute $AUM(x_i,y_i)$ for each $i$ as in Eq. (\ref{eq:aum})}
        \For {each $({x}_i,y_i) \in \mathcal{D}_{train}$}
            \If {$AUM({x}_i,y_i)<median$}
                \State {$\mathcal{D}_{low} \gets \mathcal{D}_{low} \cup ({x}_i,y_i)$}
            \ElsIf {$AUM({x}_i,y_i)\geq median$}
                \State {$\mathcal{D}_{high} \gets \mathcal{D}_{high} \cup ({x}_i,y_i)$}
            \EndIf
        \EndFor
        \State \Return $\mathcal{D}_{high}, \mathcal{D}_{low}$
    \EndFunction
\end{algorithmic}
\end{algorithm}

\begin{algorithm}[t]
\caption{: Proposed Mixup} 
\label{alg:mixup}
\begin{algorithmic}[1]
\Require \begin{varwidth}[t]{\linewidth}
$\mathcal{D}_{train}=\{({x}_i,y_i)\}_{i=1,\cdots,n}$; model $f$ 
            \end{varwidth}
    \State \multiline{$\mathcal{D}_{high}, \mathcal{D}_{low} \gets \textsc{data-categorization($\mathcal{D}_{train}$)}$  }
    \For{$k$ := 0 {to} T}
        \State $Total\_Loss \gets 0$
        \For{$i$ := 0 {to} $|\mathcal{D}_{train}|$}
            \State $Loss \gets CrossEntropy(f({x_i}),{y_i})$
            \State \multiline{Construct a saliency map $S$ by computing the gradient of $Loss$ with respect to $\textbf{z}$}
                \If{$(x_i,y_i) \in \mathcal{D}_{high}$}:
                    \State \multiline{Find the most similar/dissimilar samples from $\mathcal{D}_{low}$ using Eq. (\ref{eq:high})}
                \ElsIf{$(x_i,y_i) \in \mathcal{D}_{low}$}:
                    \State \multiline{Find the most similar/dissimilar samples from $\mathcal{D}_{high}$ using Eq. (\ref{eq:low})}
                \EndIf
            \State \multiline{Generate two mixup samples, one for $(x_i,y_i)$ and its most similar sample and another for $(x_i,y_i)$ and its most dissimilar sample, using Eq. \ref{eq:1}.}
            \State \multiline{Compute $CrossEntropy$ loss for each mixup sample}
            \State  \multiline{$Loss \gets \beta  Loss  + \gamma  Loss^{\prime} + \delta  Loss^{\prime \prime}$}
        \EndFor
        \State $Total\_Loss \gets Total\_Loss + Loss$
        \State Update the model weights
    \EndFor
\end{algorithmic}
\end{algorithm}

\paragraph{Mixup using Saliency Signals} 
We conduct a mixup operation on the two data categories generated by Algorithm \ref{alg:aum} using saliency signals (as detailed below).
For the mixup, rather than selecting random samples from $\mathcal{D}_{high}$ and $\mathcal{D}_{low}$ to mix, we utilize saliency signals to select samples. 
To measure saliency, gradient-based methods are usually used for saliency computation \cite{li-etal-2016-visualizing, rei-sogaard-2018-zero,yoon-etal-2021-ssmix}. Following this idea, we simply compute the gradient of the classification loss $L$ with respect to each logit value $z_i \in \textbf{z}$ and take the absolute value of the gradient components as the saliency map or signature $S$ for a sample $(x_i,y_i) \in \mathcal{D}_{train}$. 
For a sample $(x_i,y_i)$, we then leverage its saliency map $S$ to find the most similar and most dissimilar samples from the other data category that $(x_i,y_i)$ does not belong to according to its AUM, in order to calibrate in-domain and out-of-domain data. For example, if $(x_i,y_i) \in \mathcal{D}_{high}$, we find its most similar
sample $(x_i^\prime, y_i^\prime)$ and its most dissimilar sample $(x_i^{\prime \prime}, y_i^{\prime \prime})$ from $\mathcal{D}_{low}$, that return the largest and smallest cosine similarity, respectively, with the saliency map $S$ of $(x_i,y_i)$. That is, the most similar and most dissimilar samples to $(x_i,y_i) \in \mathcal{D}_{high}$ are calculated as follows: 
\begin{equation}
\begin{split}
    (x^{\prime}_{i},y^{\prime}_{i}) = \argmax_{(x_j,y_j)\in\mathcal{D}_{low}} CosSim (S,S^{(x_j,y_j)}) \\
    \vspace{-2mm}
     (x^{\prime \prime}_{i},y^{\prime \prime}_{i}) = \argmin_{(x_j,y_j)\in\mathcal{D}_{low}} CosSim (S,S^{(x_j,y_j)})
\end{split}
\label{eq:high}
\vspace{-2mm}
\end{equation}
Similarly, if $(x_i,y_i)$ belongs to $\mathcal{D}_{low}$, we find the most similar/dissimilar samples from $\mathcal{D}_{high}$ that return the largest/smallest cosine similarity with $S$ as follows:
\begin{equation}
\begin{split}
    (x^{\prime}_{i},y^{\prime}_{i}) = \argmax_{(x_j,y_j)\in\mathcal{D}_{high}} CosSim (S,S^{(x_j,y_j)}) \\
     (x^{\prime \prime}_{i},y^{\prime \prime}_{i}) = \argmin_{(x_j,y_j)\in\mathcal{D}_{high}} CosSim (S,S^{(x_j,y_j)})
\end{split}
\label{eq:low}
\end{equation}
We then generate two mixup samples for a given sample $(x_i,y_i)$ by interpolating the selected samples, which are the most similar sample $(x_i^\prime, y_i^\prime)$, and the most dissimilar sample $(x_i^{\prime\prime}, y_i^{\prime\prime})$. For the mixup operation, we follow the original mixing ratio sampling strategy which is shown in Eq. (\ref{eq:1}). The ratio $\lambda$ is sampled from a Beta($\alpha, \alpha$) distribution with a hyper-parameter $\alpha$. 

Intuitively, by synthesizing the original sample and the most similar sample from the other data category, we calibrate in-domain data. The augmented sample mimics in-domain sample since it aligns the most with the original sample. Furthermore, by selecting the sample from the other category, we allow the generated mixup sample to combine easy-to-learn and hard-to-learn samples properly.
By synthesizing the original and the most dissimilar sample from the other data category, we calibrate out-of-domain data. The augmented sample mimics out-of-domain instances since we pick a sample that is the most dissimilar to the original sample. As above, by selecting the sample from the other category, we allow the augmented sample to contain both information of easy-to-learn and hard-to-learn samples, useful for both optimization and learning.
Note that our mixup method mixes samples on the level of \texttt{[CLS]} hidden state representations generated by task-specific layer on top of the pre-trained language model. 
We summarize the process in Algorithm \ref{alg:mixup}. We combine each loss by weighted sum (see Alg. \ref{alg:mixup}) where $\beta, \gamma, \delta$ are hyper-parameters that sum up to 1. In our experiments, we conduct our mixup operation using mini-batch SGD to update the model weights. Note that other saliency measures are possible to compute similarity/dissimilarity between samples and will be an interesting future direction.

\subsection{Calibration Metrics}
\label{sec:ece}
A model is perfectly calibrated when the confidence estimate $\hat{p}$ of the model is equal to true probability (accuracy) $\mathbbm{P}(\hat{y}=y|\hat{p})=\hat{p}$. \cite{naeini2015obtaining, guo2017calibration,desai2020calibration}. 
This can be empirically approximated by discretizing the probability interval into a fixed number of bins $M=10$ where each bin $b_m$ contains predicted probabilities that encompass the interval. The expected calibration error (ECE) is calculated by weighting the average of the difference between each bin's accuracy and confidence as follows:

\begin{gather*}
    \mbox{acc}(b_m) =  \frac{1}{|b_m|}\sum_{i\in b_m}\mathbbm{1}(\hat{y}_i=y_i) \\ 
    \mbox{conf}(b_m) = \frac{1}{|b_m|}\sum_{i\in b_m}\hat{p_i} \\
    \mbox{ECE} = \sum_{m=1}^{M}\frac{|b_m|}{N}|\mbox{acc}(b_m)-\mbox{conf}(b_m)|
\label{eq:3}
\end{gather*}
where $N$ is the total number of predictions.

\subsection{Miscalibration Correction Methods}
We explore the combination of miscalibration correction methods (described below) with mixup to investigate their impact on our proposed mixup for model calibration. 

\paragraph{Label Smoothing (LS)}
In supervised learning, one-hot encoded labels fail to provide uncertainty of inputs due to the fact that all the probability mass is given to one class. This results in over-confident models since the largest logit becomes larger than the others which removes the uncertainty of label space. Label smoothing (LS) is a solution to penalize this by preventing the models from becoming over-confident.
In this work, we incorporate label smoothing with our proposed mixup. We generate smoothed one-hot target signal while creating mixup instances by distributing $\frac{\sigma}{|{y}|-1}$ mass over non ground-truth classes, where $\sigma \in (0,1)$ is a hyper-parameter and $|{y}|$ is the number of classes.\footnote{For example, the smoothed one-hot target of [1,0,0] is [0.99, 0.0005, 0.0005] when $\sigma=0.001$ and $|{y}|=3$.} 

\paragraph{Temperature Scaling (TS)} 
Temperature scaling (TS) is a post-processing step which re-scales the logit vector $\textbf{z}$ using a single scale parameter temperature, $T>0$ for all classes. TS has the effect of softening the outputs to be uniform with $T>1$, while $T \to 0$ has the effect of collapsing probability mass to one class. We explore the effect of TS when  incorporated with our proposed mixup.

\section{Experiments}
\subsection{Tasks and Datasets}
We evaluate our calibration-targeted mixup on three natural language understanding tasks: natural language inference, paraphrase detection, and commonsense reasoning. We evaluate the models in-domain (training and testing on data from the same distribution) and out-of-domain (training and testing on data from different distributions).
Mixup reduces the number of undesirable oscillations when predicting especially on out-of-distribution samples \cite{zhang2017mixup}.
Hence, effective mixup should be less prone to over-fitting when handling out-of-distribution data. To test the benefits of our proposed method for pre-trained language model calibration, we use in-domain trained models to predict out-of-distribution test samples. We describe our in-domain and out-of-domain sets as follows. 
\paragraph{Natural Language Inference}
Stanford Natural Language Inference (SNLI) is a natural language inference task to predict if the relation between a hypothesis and a premise is \textit{entailment, contradiction,} or \textit{neutral} \cite{bowman-etal-2015-large}.
Multi-Genre Natural Language Inference (MNLI) captures natural language inference with more diverse domains \cite{william2018mnli} than SNLI. 

\paragraph{Paraphrase Detection} 
Quora Question Pairs (QQP) is a paraphrase detection task to test if two questions are semantically equivalent \cite{lyer2017qqp}.
TwitterPPDB (TPPDB) is to determine whether sentence pairs from Twitter convey similar semantics when they share URLs \cite{lan-etal-2017-continuously}

\paragraph{Commonsense Reasoning} 
Situations With Adversarial Generations (SWAG) is a commonsense reasoning task to choose the most plausible continuation of a sentence among four candidates \cite{zellers2018swag}. 
HellaSWAG is a dataset built using adversarial filtering to generate challenging out-of-domain samples. It is distributionally different in that its examples exploit statistical biases in pre-trained models.

\subsection{Comparison Methods}
In this work, we explore the mixup effects on NLU with the goal of producing better calibrated models, in particular pre-trained language models, which are BERT \cite{devlin2019bert} and RoBERTa \cite{liu2019roberta}.
We consider the following baselines:
\begin{itemize}
    \item \textbf{Pre-trained Language Models} : Pre-trained language models fine-tuning on each downstream task using BERT \cite{devlin2019bert} and RoBERTa \cite{liu2019roberta}.
    \item \textbf{Mixup} \cite{zhang2017mixup,singh2019mixup}: Mixup augments training data by linearly interpolating randomly selected training samples in the input space. The interpolation of Mixup is performed on the input embeddings obtained from the first layer of the language model.
    \item \textbf{Manifold-mixup (M-mixup)} \cite{verma2019manifold} : An extension of Mixup, which interpolates training samples in the hidden feature space. The interpolation of Manifold-mixup is performed on the features obtained from the last layer of the language model.
\end{itemize}
Each method is compared with two variants where miscalibration correction methods (label smoothing, LS and temperature scaling, TS) are applied.\footnote{For vanila pre-trained language models with/without label smoothing results, we use the reported results from \citet{desai2020calibration}.}

\subsection{Implementation Details}
We use the same set of hyper-parameters across all tasks as \citet{desai2020calibration} for a fair comparison.
We train models with a maximum of 3 epochs. For BERT, we set batch size of 16, a learning rate of 1e-5, gradient clip of 1.0, and no weight decay. For RoBERTa, we set batch size of 32, a learning rate of 2e-5, gradient clip of 1.0, and weight decay of 0.1. We follow the published train/validation/test split by \citet{desai2020calibration}.\footnote{https://github.com/shreydesai/calibration} 
For mixup, we use a mixing ratio sampling strategy hyper-parameter $\alpha =0.4$.
We use loss weight hyper-parameters, $\beta, \gamma, \delta, $ values as $0.8/0.1/0.1$ respectively. 
We did hyper-parameter search for label smoothing $\sigma \in [0.001, 0.003, 0.01, 0.03, 0.1, 0.3]$. 
We use $\sigma=0.01/0.03/0.3$ for BERT, $\sigma=0.003/0.03/0.3$ for RoBERTa on SNLI/QQP/SWAG, respectively. 
We use threshold values for splitting data into two groups $\mathcal{D}_{high}$ and $\mathcal{D}_{low}$ (the median AUM over full training samples) as $3.5/4.4/2.5$ for BERT, $3.4/4.0/2.7$ for RoBERTa on SNLI/QQP/SWAG, respectively.
For all results, we report the mean across five training runs with random restarts. Finally, all experiments are conducted on a single NVIDIA RTX A5000 24G GPU with a total time for fine-tuning all models being under 24 hours. Temperature scaling (TS) searches are performed in the range of [0.01,5.0] with a granularity of 0.01 using development datasets. TS is completed very fast since it uses separate cached logits.

\begin{table*}[t]
\resizebox{\textwidth}{!}{%
\centering

\begin{tabular}{l|c|c|c|c|c|c|c|c|c|c|c|c}
\toprule
                        & \multicolumn{6}{c|}{In-Domain}                                                                                                                                            & \multicolumn{6}{c}{Out-of-Domain}                                                                                                                                        \\ \midrule
                      & \multicolumn{2}{c|}{SNLI}                               & \multicolumn{2}{c|}{QQP}                                & \multicolumn{2}{c|}{SWAG}                               & \multicolumn{2}{c|}{MNLI}                               & \multicolumn{2}{c|}{TwitterPPDB}                        & \multicolumn{2}{c}{HellaSWAG}                          \\ \midrule
                      & \multicolumn{1}{c|}{{No TS}} & \multicolumn{1}{c|}{{TS}} & \multicolumn{1}{c|}{{No TS}} & \multicolumn{1}{c|}{{TS}} & \multicolumn{1}{c|}{{No TS}} & \multicolumn{1}{c|}{{TS}} & \multicolumn{1}{c|}{{No TS}} & \multicolumn{1}{c|}{{TS}} & \multicolumn{1}{c|}{{No TS}} & \multicolumn{1}{c|}{{TS}} & \multicolumn{1}{c|}{{No TS}} & \multicolumn{1}{c}{{TS}} \\ \midrule
BERT                   & $2.54_{0.8}$                     & {$1.14_{1.0}$} & $2.71_{0.5}$          & $0.97_{0.1}$          & $2.49_{1.8}$          & $0.85_{0.4}$          & $7.09_{2.1}$          & $3.61_{1.7}$          & $8.51_{0.6}$           & $7.15_{0.9}$           & $12.62_{2.8}$         & $12.83_{2.1}$         \\
BERT + LS              & $7.12_{0.3}$ & $8.37_{0.5}$          & $6.33_{0.4}$          & $8.16_{0.7}$          & $10.01_{1.0}$         & $10.89_{1.1}$         & $3.74_{1.4}$          & $4.05_{0.9}$          & $6.30_{0.8}$           & $5.78_{0.7}$           & {$5.73_{0.6}$} & $5.34_{0.9}$          \\ \midrule
Mixup               & $7.73_{1.1}$                     & $3.18_{0.9}$         & $9.04_{0.8}$          & $3.36_{1.1}$          & $7.08_{1.0}$         & $2.08_{0.6}$          & $19.51_{2.1}$         & $3.56_{1.7}$          & $11.70_{1.6}$          & $5.03_{1.3}$           & $10.93_{2.0}$         &$4.24_{1.6}$           \\
Mixup + LS          & $7.92_{1.0}$                     & $2.63_{0.8}$          & $9.65_{0.4}$          & $2.49_{1.3}$          & $7.44_{0.7}$          & $1.15_{0.2}$        & $18.57_{1.2}$         & $2.31_{1.0}$          & $11.16_{0.8}$          & $4.58_{1.1}$           & $8.57_{1.3}$          & $3.95_{1.1}$              \\ 
M-Mixup       & $3.17_{0.8}$        & $1.77_{0.3}$          & $8.55_{1.2}$          & $6.11_{1.1}$          & $5.18_{0.6}$          & $1.09_{0.4}$          & $12.92_{2.6}$         &  $2.34_{1.9}$           & $12.10_{2.3}$         &  $7.98_{2.6}$         & $9.82_{1.2}$          & $5.12_{0.9}$              \\
 M-Mixup + LS & $3.40_{0.4}$                     & $5.14_{0.7}$          &  $3.49_{0.2}$  &  $3.71_{0.7}$  &  $5.24_{0.5}$  &  $1.26_{0.2}$  & $16.76_{1.3}$  &  $4.57_{0.9}$  & $6.29_{1.1}$    &  $6.54_{1.7}$   &  $8.32_{0.7}$           &  $3.64_{0.6}$  \\ \midrule
 Ours & {$\textbf{1.29}_{0.4}$}      &{$\textbf{0.77}_{0.7}$}         & {$2.05_{0.6}$} & {$1.02_{0.6}$} & {$\textbf{2.01}_{0.4}$} & {$0.81_{0.2}$} & {$2.73_{2.5}$} & {$3.50_{0.6}$} & ${5.69}_{0.7}$  & $\textbf{3.16}_{1.2}$  & ${5.49}_{1.9}$         & ${4.11}_{1.5}$ \\
Ours + LS & {$1.85_{0.3}$}                     &{$1.05_{1.0}$ }         & $\textbf{1.70}_{0.9}$ & $\textbf{0.95}_{0.1}$ & {$2.09_{0.7}$} & $\textbf{0.79}_{0.3}$ & $\textbf{2.26}_{1.0}$ & $\textbf{1.70}_{0.5}$ & $\textbf{5.37}_{1.0}$  & {$3.54_{1.1}$}  & $\textbf{4.26}_{0.8}$         & $\textbf{3.28}_{0.7}$ \\ \midrule
\toprule
                        & \multicolumn{6}{c|}{In-Domain}                                                                                                                                            & \multicolumn{6}{c}{Out-of-Domain}                                                                                                                                        \\ \midrule
                      & \multicolumn{2}{c|}{SNLI}                               & \multicolumn{2}{c|}{QQP}                                & \multicolumn{2}{c|}{SWAG}                               & \multicolumn{2}{c|}{MNLI}                               & \multicolumn{2}{c|}{TwitterPPDB}                        & \multicolumn{2}{c}{HellaSWAG}                          \\ \midrule
                    & \multicolumn{1}{c|}{{No TS}} & \multicolumn{1}{c|}{{TS}} & \multicolumn{1}{c|}{{No TS}} & \multicolumn{1}{c|}{{TS}} & \multicolumn{1}{c|}{{No TS}} & \multicolumn{1}{c|}{{TS}} & \multicolumn{1}{c|}{{No TS}} & \multicolumn{1}{c|}{{TS}} & \multicolumn{1}{c|}{{No TS}} & \multicolumn{1}{c|}{{TS}} & \multicolumn{1}{c|}{{No TS}} & \multicolumn{1}{c}{{TS}} \\ \midrule
RoBERTa               & $1.93_{0.5}$                      & $0.84_{0.8}$              & $2.33_{0.1}$                      & $0.88_{0.6}$                       & $1.76_{1.0}$                      & $\textbf{0.76}_{0.7}$                       & $3.62_{3.2}$                      & {$1.46_{2.5}$}                       & $9.55_{0.6}$                      & $7.86_{0.5}$                       & $11.93_{3.2}$                     & $11.22_{2.9}$                      \\
RoBERTa + LS           & $6.38_{0.6}$                      & $8.70_{1.0}$                       & $6.11_{0.3}$                      & $8.69_{0.6}$                       & $8.81_{0.3}$                      & $11.40_{0.6}$                      & $4.50_{1.4}$                      & $5.93_{1.9}$                       & $8.91_{0.3}$                      & $5.31_{0.7}$                       & $2.14_{1.4}$                      & $2.23_{1.1}$                       \\ \midrule
Mixup               & $7.67_{0.8}$                      & $4.51_{0.7}$                       & $3.41_{0.5}$                      & $1.64_{0.6}$                       & $3.60_{0.9}$                      & $1.03_{0.9}$                       & $16.85_{1.3}$                     & $5.65_{0.9}$                       & $11.03_{0.9}$                     & $5.41_{0.8}$                       & $7.02_{0.2}$                      & $3.90_{0.6}$                       \\
Mixup + LS          & $6.10_{0.7}$                      & $1.99_{0.5}$                       & $6.56_{0.9}$                      & $2.96_{0.5}$                       & $2.52_{0.1}$                      & $0.85_{0.3}$                       & $10.89_{1.1}$                     & $1.82_{0.3}$                       & $9.01_{1.6}$                      & $3.09_{1.1}$                       & $7.75_{1.7}$                      & $2.41_{0.7}$                       \\
M-Mixup       & $7.32_{0.8}$         & $4.56_{0.4}$          & $3.54_{0.5}$          & $5.05_{0.6}$          & $1.68_{1.2}$          & $0.96_{0.3}$          & $19.78_{3.1}$         & $7.65_{1.3}$           & $7.18_{1.8}$          &  $8.76_{2.1}$         & $5.63_{2.8}$          & $3.43_{1.5}$          \\
M-Mixup + LS & $3.51_{1.0}$                     & $3.00_{0.9}$          &  $2.82_{0.7}$  &  $3.03_{0.6}$  &  $1.83_{1.5}$  &  $0.94_{0.4}$  &  
$8.23_{1.6}$  &  $5.08_{1.0}$  &  $6.17_{0.9}$   &  $6.91_{1.1}$   &  $4.27_{0.6}$           &  $2.88_{1.6}$  \\ \midrule 
Ours & {$1.34_{0.7}$}                     & $\textbf{0.63}_{0.5}$          & {$2.47_{0.6}$}  &{$1.41_{0.2}$}  & $1.24_{0.1}$   & $1.03_{0.2}$  &{$1.41_{1.9}$}  & $\textbf{1.18}_{1.4}$ &$\textbf{3.94}_{0.9}$ & $\textbf{1.89}_{1.2}$  & $2.40_{1.8}$        & $2.08_{1.5}$ \\ 
Ours + LS & $\textbf{1.28}_{0.6}$                     & $1.02_{0.6}$          & {$\textbf{2.18}_{0.7}$}  &{$\textbf{0.84}_{0.4}$}  &{$\textbf{1.12}_{0.4}$}  &$0.81_{0.1}$  &$\textbf{1.37}_{1.7}$  & {$1.60_{1.3}$} &${3.96}_{1.6}$ & ${2.67}_{1.8}$  & $\textbf{1.86}_{0.9}$         & $\textbf{1.70}_{1.2}$
 \\ \bottomrule
\end{tabular}
}
\caption{{Expected Calibration Error (ECE) in percentage (\%) on BERT (top) and RoBERTa (bottom). Bold text shows the best ECE. Lower ECE implies better-calibrated models. We report the mean ECE across five runs with random restarts. The subscript represents the corresponding standard deviation (e.g., $1.29_{0.4}$ indicates $1.29 \pm 0.4$).}} 
\label{tb:ece_result}
\end{table*}

\begin{table*}[t]
\centering
\small
\begin{tabular}{l|ccc|ccc}
\toprule
                & \multicolumn{3}{c|}{In-Domain} & \multicolumn{3}{c}{Out-of-Domain} \\ \midrule
                & SNLI     & QQP      & SWAG    & MNLI   & TwitterPPDB  & HellaSWAG \\ \midrule
BERT        & $90.04_{0.3}$    & $90.27_{0.3}$    & $79.40_{0.4}$   & $73.52_{0.3}$  & $87.63_{0.4}$        & $34.48_{0.2}$     \\
BERT + LS        & $87.11_{0.8}$    & $87.51_{0.4}$    & $74.91_{0.3}$   & $72.06_{1.2}$  & $87.82_{0.6}$        & $36.48_{1.8}$     \\ \midrule
Mixup        & $88.82_{0.2}$    & $89.12_{0.5}$    & $74.98_{2.3}$   & $69.19_{0.8}$  & $87.45_{0.3}$        & $33.22_{0.4}$     \\ 
Mixup + LS  & $88.74_{0.4}$    & $89.24_{0.2}$    & $75.75_{0.5}$  & $69.37_{1.1}$  & $87.69_{0.6}$        & $35.65_{1.7}$     \\ 
M-Mixup      & $86.40_{0.3}$    & $89.37_{0.6}$    & $76.96_{0.4}$   & $66.61_{0.6}$ & $86.51_{0.8}$        & $34.57_{1.4}$     \\
M-Mixup + LS & $87.50_{0.7}$    & $87.17_{0.6}$    & $76.09_{0.9}$   & $64.88_{0.9}$  & $86.55_{1.1}$        & $33.71_{0.6}$     \\ \midrule
Ours  &$90.01_{0.4}$ &$90.13_{0.2}$ &$78.94_{0.8}$ & $73.48_{0.4}$ &$88.04_{0.7}$ &$34.63_{0.4}$ \\
Ours + LS  &$90.14_{0.3}$ &$90.32_{0.2}$ &$79.26_{0.6}$ & $72.36_{0.6}$ &$87.62_{0.9}$ &$34.97_{0.5}$ \\
\bottomrule
\toprule
                & \multicolumn{3}{c|}{In-Domain} & \multicolumn{3}{c}{Out-of-Domain} \\ \midrule
                & SNLI     & QQP      & SWAG    & MNLI   & TwitterPPDB  & HellaSWAG \\ \midrule
RoBERTa     & $91.23_{0.3}$    & $91.11_{0.2}$    & $82.45_{1.2}$   & $78.79_{0.2}$  & $86.72_{0.2}$        & $41.68_{1.1}$     \\
RoBERTa + LS    & $89.73_{0.4}$    & $87.64_{0.4}$    & $79.13_{0.4}$   & $77.40_{0.5}$  & $87.48_{1.2}$        & $40.05_{0.9}$     \\ \midrule
Mixup        & $90.59_{0.4}$    & $89.20_{1.4}$    & $79.91_{1.5}$   & $75.74_{0.7}$  & $84.74_{0.6}$        & $40.92_{1.4}$     \\
Mixup + LS   & $90.44_{0.6}$    & $87.45_{0.7}$    & $79.16_{0.4}$   & $76.44_{1.0}$  & $87.48_{0.4}$        & $39.95_{1.0}$     \\
M-Mixup      & $90.30_{0.5}$    & $89.47_{0.7}$    & $73.79_{0.8}$   & $73.69_{1.0}$  & $86.04_{0.7}$        & $41.60_{0.8}$     \\
M-Mixup + LS & $90.97_{0.4}$    & $88.44_{1.0}$    & $79.61_{0.6}$   & $75.55_{0.9}$  & $86.49_{1.5}$        & $41.88_{1.1}$    \\ \midrule
Ours  & $91.61_{0.5}$ &$89.19_{0.4}$ &$81.47_{0.8}$ & $78.01_{0.6}$ & $87.13_{0.8}$  & $40.95_{1.4}$ \\
Ours + LS & $91.24_{0.3}$ &$89.75_{0.6}$ & $82.69_{0.7}$ & $78.86_{0.5}$ & $87.63_{1.0}$  & $41.37_{1.1}$ \\
\bottomrule
\end{tabular}
\caption{{The comparison of accuracy (\%) on BERT (top) and RoBERTa (bottom). We report the mean accuracy across five training runs with the standard deviation shown in subscript (e.g., $90.01_{0.4}$ indicates $90.01 \pm 0.4$).}}
\label{tb:accuracy}
\end{table*}

\begin{table*}[t]
\resizebox{\textwidth}{!}{%
\begin{tabular}{llcccccccccccc}
\toprule
                                              & \multicolumn{1}{l|}{}             & \multicolumn{6}{c|}{In-Domain}                                                                                                                                                 & \multicolumn{6}{c}{Out-of-domain}                                                                                                                         \\ \hline
                                              & \multicolumn{1}{l|}{}             & \multicolumn{2}{c|}{SNLI}                                & \multicolumn{2}{c|}{QQP}                                 & \multicolumn{2}{c|}{SWAG}                                & \multicolumn{2}{c|}{MNLI}                                & \multicolumn{2}{c|}{TwitterPPDB}                         & \multicolumn{2}{c}{HellaSWAG}       \\ \hline
                                              & \multicolumn{1}{l|}{}             & \multicolumn{1}{c|}{No TS} & \multicolumn{1}{c|}{TS} & \multicolumn{1}{c|}{No TS} & \multicolumn{1}{c|}{TS} & \multicolumn{1}{c|}{No TS} & \multicolumn{1}{c|}{TS} & \multicolumn{1}{c|}{No TS} & \multicolumn{1}{c|}{TS} & \multicolumn{1}{c|}{No TS} & \multicolumn{1}{c|}{TS} & \multicolumn{1}{c|}{No TS} & TS \\ \hline
\multicolumn{1}{l|}{\multirow{5}{*}{\rotatebox[]{90}{BERT}}}    & \multicolumn{1}{l|}{Ours}         & \multicolumn{1}{c|}{1.85}  & \multicolumn{1}{c|}{1.05}   & \multicolumn{1}{c|}{1.70}  & \multicolumn{1}{c|}{0.95}   & \multicolumn{1}{c|}{2.09}  & \multicolumn{1}{c|}{0.79}   & \multicolumn{1}{c|}{2.26}  & \multicolumn{1}{c|}{1.70}   & \multicolumn{1}{c|}{5.37}  & \multicolumn{1}{c|}{3.54}   & \multicolumn{1}{c|}{4.26}  & 3.28   \\ \cline{2-14} 
\multicolumn{1}{l|}{}                         & \multicolumn{1}{l|}{- AUM}      & \multicolumn{1}{c|}{2.74}  & \multicolumn{1}{c|}{0.95}   & \multicolumn{1}{c|}{4.43}  & \multicolumn{1}{c|}{1.39}   & \multicolumn{1}{c|}{2.15}  & \multicolumn{1}{c|}{1.44}   & \multicolumn{1}{c|}{7.74}  & \multicolumn{1}{c|}{1.68}   & \multicolumn{1}{c|}{9.08}  & \multicolumn{1}{c|}{4.17}   & \multicolumn{1}{c|}{11.45} & 2.21   \\ 
\multicolumn{1}{l|}{}                         & \multicolumn{1}{l|}{- Saliency} & \multicolumn{1}{c|}{2.34}  & \multicolumn{1}{c|}{3.16}   & \multicolumn{1}{c|}{5.97}  & \multicolumn{1}{c|}{4.94}   & \multicolumn{1}{c|}{4.19}  & \multicolumn{1}{c|}{1.11}   & \multicolumn{1}{c|}{9.51}  & \multicolumn{1}{c|}{4.14}   & \multicolumn{1}{c|}{6.03}  & \multicolumn{1}{c|}{6.79}   & \multicolumn{1}{c|}{7.91}  & 4.28   \\ \cline{2-14}
\multicolumn{1}{l|}{}    & \multicolumn{1}{l|}{- dissimilar} & \multicolumn{1}{c|}{0.60}  & \multicolumn{1}{c|}{0.76}   & \multicolumn{1}{c|}{1.51}  & \multicolumn{1}{c|}{1.16}   & \multicolumn{1}{c|}{4.33}  & \multicolumn{1}{c|}{0.81}   & \multicolumn{1}{c|}{4.91}  & \multicolumn{1}{c|}{2.52}   & \multicolumn{1}{c|}{8.33}  & \multicolumn{1}{c|}{4.32}   & \multicolumn{1}{c|}{12.60} & 6.38   \\
\multicolumn{1}{l|}{}                         & \multicolumn{1}{l|}{- similar}    & \multicolumn{1}{c|}{3.76}  & \multicolumn{1}{c|}{4.94}   & \multicolumn{1}{c|}{2.88}  & \multicolumn{1}{c|}{1.98}   & \multicolumn{1}{c|}{5.58}  & \multicolumn{1}{c|}{2.87}        & \multicolumn{1}{c|}{8.38}  & \multicolumn{1}{c|}{3.07}   & \multicolumn{1}{c|}{7.67}  & \multicolumn{1}{c|}{5.55}   & \multicolumn{1}{c|}{18.91} & 3.24   \\ \hline
                                              &                                   &                            &                             &                            &                             &                            &                             &                            &                             &                            &                             &                            &        \\ \hline
\multicolumn{1}{l|}{\multirow{5}{*}{\rotatebox[]{90}{RoBERTa}}} & \multicolumn{1}{l|}{Ours}         & \multicolumn{1}{c|}{1.28}  & \multicolumn{1}{c|}{1.08}   & \multicolumn{1}{c|}{2.18}  & \multicolumn{1}{c|}{0.84}   & \multicolumn{1}{c|}{1.12}  & \multicolumn{1}{c|}{0.81}   & \multicolumn{1}{c|}{1.37}  & \multicolumn{1}{c|}{1.60}   & \multicolumn{1}{c|}{3.96}  & \multicolumn{1}{c|}{2.67}   & \multicolumn{1}{c|}{1.86}  & 1.70   \\ \cline{2-14} 
\multicolumn{1}{l|}{}                         & \multicolumn{1}{l|}{- AUM}      & \multicolumn{1}{c|}{5.18}  & \multicolumn{1}{c|}{2.25}   & \multicolumn{1}{c|}{3.59}  & \multicolumn{1}{c|}{0.79}   & \multicolumn{1}{c|}{2.31}  & \multicolumn{1}{c|}{1.39}   & \multicolumn{1}{c|}{11.29} & \multicolumn{1}{c|}{5.75}   & \multicolumn{1}{c|}{8.09}  & \multicolumn{1}{c|}{1.78}   & \multicolumn{1}{c|}{12.46} & 3.79   \\
\multicolumn{1}{l|}{}                         & \multicolumn{1}{l|}{- Saliency} & \multicolumn{1}{c|}{2.91}  & \multicolumn{1}{c|}{2.63}   & \multicolumn{1}{c|}{0.98}  & \multicolumn{1}{c|}{1.02}   & \multicolumn{1}{c|}{1.41}  & \multicolumn{1}{c|}{1.27}   & \multicolumn{1}{c|}{4.80}  & \multicolumn{1}{c|}{4.54}   & \multicolumn{1}{c|}{6.92}  & \multicolumn{1}{c|}{4.78}   & \multicolumn{1}{c|}{6.82}  & 3.37   \\ \cline{2-14}
\multicolumn{1}{l|}{} & \multicolumn{1}{l|}{- dissimilar} & \multicolumn{1}{c|}{2.01}  & \multicolumn{1}{c|}{0.93}   & \multicolumn{1}{c|}{2.98}  & \multicolumn{1}{c|}{1.58}   & \multicolumn{1}{c|}{2.52}  & \multicolumn{1}{c|}{0.73}   & \multicolumn{1}{c|}{6.69}  & \multicolumn{1}{c|}{4.77}   & \multicolumn{1}{c|}{5.17}  & \multicolumn{1}{c|}{4.54}   & \multicolumn{1}{c|}{11.39} & 6.43   \\
\multicolumn{1}{l|}{}                         &\multicolumn{1}{l|}{- similar}    & \multicolumn{1}{c|}{2.69}      & \multicolumn{1}{c|}{2.33}       & \multicolumn{1}{c|}{5.14}  & \multicolumn{1}{c|}{3.40}   & \multicolumn{1}{c|}{3.10}  & \multicolumn{1}{c|}{2.62}        & \multicolumn{1}{c|}{2.01}      & \multicolumn{1}{c|}{1.84}       & \multicolumn{1}{c|}{10.43} & \multicolumn{1}{c|}{8.94}   & \multicolumn{1}{c|}{7.87}  & 6.11   \\ \bottomrule
\end{tabular}%
}
\caption{{Ablation study to investigate the effect of each component in our proposed mixup. We report results (\% ECE) of our mixup without using AUM (i.e., -AUM), without using saliency (i.e., -Saliency), without utilizing the most dissimilar sample selected from the other data category obtained by AUM (i.e., -dissimilar), and without utilizing the most similar sample selected from the other data category obtained by AUM (i.e., -similar). }}
\label{tb:ab}
\end{table*}

\subsection{Results}
We show the comparison of experimental results (ECE) on BERT and RoBERTa in Table \ref{tb:ece_result}. For each task, we train the model on in-domain training set, and evaluate its expected calibration errors (ECEs) on in-domain and out-of-domain test sets.
We make the following observations:

First, for in-domain data, label smoothing (LS) does not exhibit its effectiveness on pre-trained language models' calibration. 
Specifically, for in-domain data, pre-trained language models with LS (i.e., BERT+LS/RoBERTa+LS) achieve higher expected calibration errors (ECEs) compared with vanilla pre-trained language models (i.e., BERT/RoBERTa) on all tasks. 
In contrast, out-of-domain gains benefit from LS (except RoBERTa on MNLI). 
From these results, we conclude that simply incorporating label uncertainty (through label smoothing) is not an effective regularization method since LS does not consistently improve the model calibration (especially for the in-domain setting). 
While temperature scaling (TS) corrects the miscalibration of vanilla pre-trained language models (see BERT/RoBERTa No TS vs. TS in the table), it fails to correct miscalibrated pre-trained language models with LS (see BERT+LS/RoBERTa+LS No TS vs. TS) in-domain. 
Interestingly, for some cases of out-of-domain data, pre-trained language models with LS show comparatively low ECEs while TS further reduces ECEs (e.g., BERT(LS) on TwitterPPDB/HellaSWAG, RoBERTa(LS) on TwitterPPDB). However, its impact is not enough as it still results in high ECE. This implies that TS is not a notable strategy either to pre-trained language models' calibration. Accordingly, we conclude that stronger regularization techniques are required to calibrate the pre-trained language models. 

Second, we find that mixup on the hidden feature space (i.e., M-Mixup) generally yields lower ECE than mixup on the input embedding space (i.e., Mixup) on most tasks. We infer that Mixup generates augmented samples that are not ``good'' for model calibration (i.e., semantically or syntactically) and fails to encourage regularization effects that arise from mixup. We observe that mixup training with LS is beneficial to reduce ECEs on some tasks. 
We find that TS leads to much lower ECEs on Mixup and M-Mixup (with and without LS) on most tasks. 
However, this implies that baseline mixup methods fail to produce well-calibrated models independently (without LS or TS). 
This supports our intuition and motivation for the design of a more robust approach of mixup. 

Third, we observe that our proposed mixup yields the best calibrated models (lowest ECEs) both on in-domain and out-of-domain data (except on SWAG with RoBERTa). 
We observe that often LS effectively operates along with our proposed mixup and achieves the lowest ECEs on most tasks on in-domain and out-of-domain settings.
In contrast to baseline mixup methods, our proposed mixup performs well on in-domain and out-of-domain even without applying post-calibration correction TS (see ECE values of baselines compared with our ECE values).
We also observe that TS improves the model calibration further on our mixup training in most cases.
Accordingly, we confirm the robustness of our AUM and saliency guided mixup for pre-trained language models calibration.

\paragraph{Accuracy}
We explore the accuracy of mixup training and show comparisons in Table \ref{tb:accuracy}. 
We make the following observations:
1) Both BERT+LS/RoBERTa+LS generally lead to substantial accuracy drops especially on in-domain compared with BERT/RoBERTa (i.e., 4.49\% accuracy drops on SWAG). This implies that label smoothing (LS) fails to improve model generalization by simply manipulating labels (changing from hard to soft labels).
This potentially leads to a loss of information that is correlated to model generalization \cite{muller2019does}. 2) Mixup and M-Mixup fail to achieve an accuracy that is as good as that of vanilla pre-trained language models, potentially due to an increased chance of manifold intrusion resulting from conflicts between the synthetic samples of the mixup and original training data \cite{guo2019mixup}. 3) In contrast, our proposed mixup method generally achieves competitive accuracy regardless of applying LS or not.
This evidence supports the robustness of our proposed mixup. 
Note that TS does not affect the model's accuracy because it does not change the maximum of the softmax function. 

\subsection{Ablation Study}
\paragraph{Effect of AUM and Saliency} We investigate the effectiveness of each component (i.e., AUM and saliency) in our proposed mixup. 
As shown in Table \ref{tb:ab}, our proposed mixup without the AUM (i.e., -AUM) and without saliency (i.e., -Saliency) generally increase the expected calibration errors.
In our method without using AUM, we randomly divide training data into two categories and conduct mixup operation based on saliency map. In our method without using saliency, we randomly pick two samples from the opposite low and high AUM set and conduct mixup operation. 
The results demonstrate that both metrics (AUM and saliency) are required to improve model calibration.

\paragraph{Effect of selecting the most similar and dissimilar samples} We explore the effectiveness of selecting the most similar and dissimilar samples, which are used for mixing purposes for in-domain and out-of-domain calibration, respectively. Specifically, in our proposed mixup, we synthesize additional samples that mimic in-domain data by selecting the most similar sample from the other category (e.g., an easy-to-learn sample is mixed with a hard-to-learn/ambiguous sample that is most similar to the easy-to-learn sample, by saliency maps). This is because the selected sample aligns the most with the given sample.
This intuitively results in better model generalization due to the effect arising from data augmentation (i.e., augmenting samples that are particularly similar to in-domain data) and allows better in-domain calibration. Similarly, we calibrate out-of-domain by augmenting a sample that mimics out-of-domain distribution. This is because we select the sample that is the most different from a given sample by selecting the most dissimilar sample from the other category. To verify this intuition, we conduct our proposed mixup when excluding the most similar instance (i.e., -similar) and the most dissimilar instance (i.e., -dissimilar), respectively. 

Table \ref{tb:ab} shows the results of this ablation. 
We observe that our proposed mixup without using the most dissimilar sample (i.e., -dissimilar) results in higher ECEs compared with our mixup that uses dissimilar samples on all tasks in the out-of-domain setting for both BERT and RoBERTa. 
Interestingly, we observe that our proposed mixup without using the most similar sample (i.e., -similar) results in higher ECEs compared with our mixup that uses the most similar samples on in-domain and out-of-domain data for both BERT and RoBERTa. 
These results support that selecting the most similar/dissimilar samples effectively calibrates pre-trained models for in-domain/out-of-domain data.

\vspace{-2mm}
\section{Conclusion}
We proposed a novel mixup guided by the Area Under the Margins (AUM) and saliency maps to mitigate the miscalibration of pre-trained language models BERT and RoBERTa. 
We showed that our proposed mixup method achieves the lowest Expected Calibration Errors (ECEs) for both pre-trained language models on various types of natural language understanding tasks, for both in-domain and out-of-domain data. 
For future work, we will enhance our proposed mixup further, focusing not only on model calibration but also on performance gains.
Exploring different saliency maps for computing sample similarity/disimilarity (and its degree) is another interesting future direction.

\section*{Acknowledgements}  This research is supported in part by NSF CAREER award \#1802358 and NSF CRI award \#1823292. Any opinions, findings, and conclusions expressed here are those of the authors and do not necessarily reflect the views of NSF. We thank AWS for computing resources. We also thank our anonymous reviewers for their constructive feedback and comments, which helped improve our paper.

\bibliography{acl2022}

\begin{thebibliography}{37}
\expandafter\ifx\csname natexlab\endcsname\relax\def\natexlab#1{#1}\fi

\bibitem[{Bowman et~al.(2015)Bowman, Angeli, Potts, and
  Manning}]{bowman-etal-2015-large}
Samuel~R. Bowman, Gabor Angeli, Christopher Potts, and Christopher~D. Manning.
  2015.
\newblock \href {https://doi.org/10.18653/v1/D15-1075} {A large annotated
  corpus for learning natural language inference}.
\newblock In \emph{Proceedings of the 2015 Conference on Empirical Methods in
  Natural Language Processing}, pages 632--642, Lisbon, Portugal. Association
  for Computational Linguistics.

\bibitem[{Chen et~al.(2020)Chen, Yang, and Yang}]{chen-etal-2020-mixtext}
Jiaao Chen, Zichao Yang, and Diyi Yang. 2020.
\newblock \href {https://doi.org/10.18653/v1/2020.acl-main.194} {{M}ix{T}ext:
  Linguistically-informed interpolation of hidden space for semi-supervised
  text classification}.
\newblock In \emph{Proceedings of the 58th Annual Meeting of the Association
  for Computational Linguistics}, pages 2147--2157, Online. Association for
  Computational Linguistics.

\bibitem[{Desai and Durrett(2020)}]{desai2020calibration}
Shrey Desai and Greg Durrett. 2020.
\newblock \href {https://arxiv.org/abs/2003.07892} {Calibration of pre-trained
  transformers}.
\newblock In \emph{Proceedings of the 2020 Conference on Empirical Methods in
  Natural Language Processing (EMNLP)}, pages 295--302.

\bibitem[{Devlin et~al.(2019)Devlin, Chang, Lee, and
  Toutanova}]{devlin2019bert}
Jacob Devlin, Ming-Wei Chang, Kenton Lee, and Kristina Toutanova. 2019.
\newblock \href {https://arxiv.org/abs/1810.04805} {Bert: Pre-training of deep
  bidirectional transformers for language understanding}.
\newblock In \emph{Proceedings of the 2019 Conference of the North American
  Chapter of the Association for Computational Linguistics: Human Language
  Technologies, Volume 1 (Long and Short Papers)}, pages 4171--4186.

\bibitem[{Guo et~al.(2017)Guo, Pleiss, Sun, and
  Weinberger}]{guo2017calibration}
Chuan Guo, Geoff Pleiss, Yu~Sun, and Kilian~Q Weinberger. 2017.
\newblock \href {https://arxiv.org/abs/1706.04599} {On calibration of modern
  neural networks}.
\newblock In \emph{International Conference on Machine Learning}, pages
  1321--1330. PMLR.

\bibitem[{Guo et~al.(2019{\natexlab{a}})Guo, Mao, and
  Zhang}]{guo2019augmenting}
Hongyu Guo, Yongyi Mao, and Richong Zhang. 2019{\natexlab{a}}.
\newblock \href {https://arxiv.org/abs/1905.08941} {Augmenting data with mixup
  for sentence classification: An empirical study}.
\newblock \emph{arXiv preprint arXiv:1905.08941}.

\bibitem[{Guo et~al.(2019{\natexlab{b}})Guo, Mao, and Zhang}]{guo2019mixup}
Hongyu Guo, Yongyi Mao, and Richong Zhang. 2019{\natexlab{b}}.
\newblock Mixup as locally linear out-of-manifold regularization.
\newblock In \emph{Proceedings of the AAAI Conference on Artificial
  Intelligence}, 01, pages 3714--3722.

\bibitem[{He et~al.(2021)He, McCann, Xiong, and
  Hosseini-Asl}]{he-etal-2021-joint}
Tianxing He, Bryan McCann, Caiming Xiong, and Ehsan Hosseini-Asl. 2021.
\newblock \href {https://aclanthology.org/2021.eacl-main.151} {Joint
  energy-based model training for better calibrated natural language
  understanding models}.
\newblock In \emph{Proceedings of the 16th Conference of the European Chapter
  of the Association for Computational Linguistics: Main Volume}, pages
  1754--1761, Online. Association for Computational Linguistics.

\bibitem[{Iyer et~al.(2017)Iyer, Dandekar, and Csernai}]{lyer2017qqp}
Shankar Iyer, Nikhil Dandekar, and Kornel Csernai. 2017.
\newblock Quora question pairs.
\newblock In \emph{First Quora Dataset Release: Question Pairs}.

\bibitem[{Jagannatha and Yu(2020)}]{jagannatha-yu-2020-calibrating}
Abhyuday Jagannatha and Hong Yu. 2020.
\newblock \href {https://doi.org/10.18653/v1/2020.acl-main.188} {Calibrating
  structured output predictors for natural language processing}.
\newblock In \emph{Proceedings of the 58th Annual Meeting of the Association
  for Computational Linguistics}, pages 2078--2092, Online. Association for
  Computational Linguistics.

\bibitem[{Jung et~al.(2020)Jung, Kang, Cheng, Mentch, and
  Schaaf}]{jung-etal-2020-posterior}
Taehee Jung, Dongyeop Kang, Hua Cheng, Lucas Mentch, and Thomas Schaaf. 2020.
\newblock \href {https://doi.org/10.18653/v1/2020.acl-main.242} {Posterior
  calibrated training on sentence classification tasks}.
\newblock In \emph{Proceedings of the 58th Annual Meeting of the Association
  for Computational Linguistics}, pages 2723--2730, Online. Association for
  Computational Linguistics.

\bibitem[{Kong et~al.(2020)Kong, Jiang, Zhuang, Lyu, Zhao, and
  Zhang}]{kong-etal-2020-calibrated}
Lingkai Kong, Haoming Jiang, Yuchen Zhuang, Jie Lyu, Tuo Zhao, and Chao Zhang.
  2020.
\newblock \href {https://doi.org/10.18653/v1/2020.emnlp-main.102} {Calibrated
  language model fine-tuning for in- and out-of-distribution data}.
\newblock In \emph{Proceedings of the 2020 Conference on Empirical Methods in
  Natural Language Processing (EMNLP)}, pages 1326--1340, Online. Association
  for Computational Linguistics.

\bibitem[{Kumar and Sarawagi(2019)}]{kumar2019calibration}
Aviral Kumar and Sunita Sarawagi. 2019.
\newblock \href {https://arxiv.org/abs/1903.00802} {Calibration of encoder
  decoder models for neural machine translation}.
\newblock \emph{arXiv preprint arXiv:1903.00802}.

\bibitem[{Lan et~al.(2017)Lan, Qiu, He, and Xu}]{lan-etal-2017-continuously}
Wuwei Lan, Siyu Qiu, Hua He, and Wei Xu. 2017.
\newblock \href {https://doi.org/10.18653/v1/D17-1126} {A continuously growing
  dataset of sentential paraphrases}.
\newblock In \emph{Proceedings of the 2017 Conference on Empirical Methods in
  Natural Language Processing}, pages 1224--1234, Copenhagen, Denmark.
  Association for Computational Linguistics.

\bibitem[{Li et~al.(2019)Li, Jin, Liu, Rawat, Cai, and
  Yu}]{info:doi/10.2196/14830}
Fei Li, Yonghao Jin, Weisong Liu, Bhanu Pratap~Singh Rawat, Pengshan Cai, and
  Hong Yu. 2019.
\newblock \href {https://doi.org/10.2196/14830} {Fine-tuning bidirectional
  encoder representations from transformers (bert)--based models on large-scale
  electronic health record notes: An empirical study}.
\newblock \emph{JMIR Med Inform}, 7(3):e14830.

\bibitem[{Li et~al.(2016)Li, Chen, Hovy, and
  Jurafsky}]{li-etal-2016-visualizing}
Jiwei Li, Xinlei Chen, Eduard Hovy, and Dan Jurafsky. 2016.
\newblock \href {https://doi.org/10.18653/v1/N16-1082} {Visualizing and
  understanding neural models in {NLP}}.
\newblock In \emph{Proceedings of the 2016 Conference of the North {A}merican
  Chapter of the Association for Computational Linguistics: Human Language
  Technologies}, pages 681--691, San Diego, California. Association for
  Computational Linguistics.

\bibitem[{Liu et~al.(2019)Liu, Ott, Goyal, Du, Joshi, Chen, Levy, Lewis,
  Zettlemoyer, and Stoyanov}]{liu2019roberta}
Yinhan Liu, Myle Ott, Naman Goyal, Jingfei Du, Mandar Joshi, Danqi Chen, Omer
  Levy, Mike Lewis, Luke Zettlemoyer, and Veselin Stoyanov. 2019.
\newblock \href {https://arxiv.org/abs/1907.11692} {Roberta: A robustly
  optimized bert pretraining approach}.
\newblock \emph{arXiv preprint arXiv:1907.11692}.

\bibitem[{M{\"u}ller et~al.(2019)M{\"u}ller, Kornblith, and
  Hinton}]{muller2019does}
Rafael M{\"u}ller, Simon Kornblith, and Geoffrey~E Hinton. 2019.
\newblock \href {https://arxiv.org/abs/1906.02629} {When does label smoothing
  help?}
\newblock \emph{Advances in Neural Information Processing Systems},
  32:4694--4703.

\bibitem[{Naeini et~al.(2015)Naeini, Cooper, and
  Hauskrecht}]{naeini2015obtaining}
Mahdi~Pakdaman Naeini, Gregory Cooper, and Milos Hauskrecht. 2015.
\newblock \href {https://dl.acm.org/doi/10.5555/2888116.2888120} {Obtaining
  well calibrated probabilities using bayesian binning}.
\newblock In \emph{Proceedings of the AAAI Conference on Artificial
  Intelligence}, volume~29.

\bibitem[{Nguyen and O{'}Connor(2015)}]{nguyen-oconnor-2015-posterior}
Khanh Nguyen and Brendan O{'}Connor. 2015.
\newblock \href {https://doi.org/10.18653/v1/D15-1182} {Posterior calibration
  and exploratory analysis for natural language processing models}.
\newblock In \emph{Proceedings of the 2015 Conference on Empirical Methods in
  Natural Language Processing}, pages 1587--1598, Lisbon, Portugal. Association
  for Computational Linguistics.

\bibitem[{Pleiss et~al.(2020)Pleiss, Zhang, Elenberg, and
  Weinberger}]{NEURIPS2020_c6102b37}
Geoff Pleiss, Tianyi Zhang, Ethan Elenberg, and Kilian~Q Weinberger. 2020.
\newblock \href
  {https://proceedings.neurips.cc/paper/2020/file/c6102b3727b2a7d8b1bb6981147081ef-Paper.pdf}
  {Identifying mislabeled data using the area under the margin ranking}.
\newblock In \emph{Advances in Neural Information Processing Systems},
  volume~33, pages 17044--17056. Curran Associates, Inc.

\bibitem[{Rei and S{\o}gaard(2018)}]{rei-sogaard-2018-zero}
Marek Rei and Anders S{\o}gaard. 2018.
\newblock \href {https://doi.org/10.18653/v1/N18-1027} {Zero-shot sequence
  labeling: Transferring knowledge from sentences to tokens}.
\newblock In \emph{Proceedings of the 2018 Conference of the North {A}merican
  Chapter of the Association for Computational Linguistics: Human Language
  Technologies, Volume 1 (Long Papers)}, pages 293--302, New Orleans,
  Louisiana. Association for Computational Linguistics.

\bibitem[{Sarabadani(2019)}]{sarabadani-2019-detection}
Sarah Sarabadani. 2019.
\newblock \href {https://doi.org/10.18653/v1/W19-3221} {Detection of adverse
  drug reaction mentions in tweets using {ELM}o}.
\newblock In \emph{Proceedings of the Fourth Social Media Mining for Health
  Applications ({\#}SMM4H) Workshop {\&} Shared Task}, pages 120--122,
  Florence, Italy. Association for Computational Linguistics.

\bibitem[{Simonyan et~al.(2013)Simonyan, Vedaldi, and
  Zisserman}]{simonyan2013deep}
Karen Simonyan, Andrea Vedaldi, and Andrew Zisserman. 2013.
\newblock \href {https://arxiv.org/abs/1312.6034} {Deep inside convolutional
  networks: Visualising image classification models and saliency maps}.
\newblock \emph{arXiv preprint arXiv:1312.6034}.

\bibitem[{Sun et~al.(2020)Sun, Xia, Yin, Liang, Yu, and
  He}]{sun-etal-2020-mixup}
Lichao Sun, Congying Xia, Wenpeng Yin, Tingting Liang, Philip Yu, and Lifang
  He. 2020.
\newblock \href {https://doi.org/10.18653/v1/2020.coling-main.305}
  {Mixup-transformer: Dynamic data augmentation for {NLP} tasks}.
\newblock In \emph{Proceedings of the 28th International Conference on
  Computational Linguistics}, pages 3436--3440, Barcelona, Spain (Online).
  International Committee on Computational Linguistics.

\bibitem[{Swayamdipta et~al.(2020)Swayamdipta, Schwartz, Lourie, Wang,
  Hajishirzi, Smith, and Choi}]{swayamdipta2020dataset}
Swabha Swayamdipta, Roy Schwartz, Nicholas Lourie, Yizhong Wang, Hannaneh
  Hajishirzi, Noah~A Smith, and Yejin Choi. 2020.
\newblock \href {https://arxiv.org/abs/2009.10795} {Dataset cartography:
  Mapping and diagnosing datasets with training dynamics}.
\newblock In \emph{Proceedings of the 2020 Conference on Empirical Methods in
  Natural Language Processing (EMNLP)}, pages 9275--9293.

\bibitem[{Thulasidasan et~al.(2019)Thulasidasan, Chennupati, Bilmes,
  Bhattacharya, and Michalak}]{singh2019mixup}
Sunil Thulasidasan, Gopinath Chennupati, Jeff~A Bilmes, Tanmoy Bhattacharya,
  and Sarah Michalak. 2019.
\newblock \href
  {https://proceedings.neurips.cc/paper/2019/file/36ad8b5f42db492827016448975cc22d-Paper.pdf}
  {On mixup training: Improved calibration and predictive uncertainty for deep
  neural networks}.
\newblock In \emph{Advances in Neural Information Processing Systems},
  volume~32. Curran Associates, Inc.

\bibitem[{Verma et~al.(2019)Verma, Lamb, Beckham, Najafi, Mitliagkas,
  Lopez-Paz, and Bengio}]{verma2019manifold}
Vikas Verma, Alex Lamb, Christopher Beckham, Amir Najafi, Ioannis Mitliagkas,
  David Lopez-Paz, and Yoshua Bengio. 2019.
\newblock \href {https://arxiv.org/abs/1806.05236} {Manifold mixup: Better
  representations by interpolating hidden states}.
\newblock In \emph{International Conference on Machine Learning}, pages
  6438--6447. PMLR.

\bibitem[{Wang et~al.(2020{\natexlab{a}})Wang, Tu, Shi, and
  Liu}]{wang-etal-2020-inference}
Shuo Wang, Zhaopeng Tu, Shuming Shi, and Yang Liu. 2020{\natexlab{a}}.
\newblock \href {https://doi.org/10.18653/v1/2020.acl-main.278} {On the
  inference calibration of neural machine translation}.
\newblock In \emph{Proceedings of the 58th Annual Meeting of the Association
  for Computational Linguistics}, pages 3070--3079, Online. Association for
  Computational Linguistics.

\bibitem[{Wang et~al.(2020{\natexlab{b}})Wang, Tu, Shi, and
  Liu}]{wang2020inference}
Shuo Wang, Zhaopeng Tu, Shuming Shi, and Yang Liu. 2020{\natexlab{b}}.
\newblock \href {https://arxiv.org/abs/2005.00963} {On the inference
  calibration of neural machine translation}.
\newblock In \emph{Proceedings of the 58th Annual Meeting of the Association
  for Computational Linguistics}, pages 3070--3079.

\bibitem[{Williams et~al.(2018)Williams, Nangia, and Bowman}]{william2018mnli}
Adina Williams, Nikita Nangia, and Samuel Bowman. 2018.
\newblock \href {http://aclweb.org/anthology/N18-1101} {A broad-coverage
  challenge corpus for sentence understanding through inference}.
\newblock In \emph{Proceedings of the 2018 Conference of the North American
  Chapter of the Association for Computational Linguistics: Human Language
  Technologies, Volume 1 (Long Papers)}, pages 1112--1122. Association for
  Computational Linguistics.

\bibitem[{Yin et~al.(2021)Yin, Wang, Qu, and Xiong}]{yin-etal-2021-batchmixup}
Wenpeng Yin, Huan Wang, Jin Qu, and Caiming Xiong. 2021.
\newblock \href {https://doi.org/10.18653/v1/2021.findings-acl.434}
  {{B}atch{M}ixup: Improving training by interpolating hidden states of the
  entire mini-batch}.
\newblock In \emph{Findings of the Association for Computational Linguistics:
  ACL-IJCNLP 2021}, pages 4908--4912, Online. Association for Computational
  Linguistics.

\bibitem[{Yoon et~al.(2021)Yoon, Kim, and Park}]{yoon-etal-2021-ssmix}
Soyoung Yoon, Gyuwan Kim, and Kyumin Park. 2021.
\newblock \href {https://doi.org/10.18653/v1/2021.findings-acl.285} {{SSM}ix:
  Saliency-based span mixup for text classification}.
\newblock In \emph{Findings of the Association for Computational Linguistics:
  ACL-IJCNLP 2021}, pages 3225--3234, Online. Association for Computational
  Linguistics.

\bibitem[{Zellers et~al.(2018)Zellers, Bisk, Schwartz, and
  Choi}]{zellers2018swag}
Rowan Zellers, Yonatan Bisk, Roy Schwartz, and Yejin Choi. 2018.
\newblock \href {https://arxiv.org/abs/1808.05326} {Swag: A large-scale
  adversarial dataset for grounded commonsense inference}.
\newblock In \emph{Proceedings of the 2018 Conference on Empirical Methods in
  Natural Language Processing}, pages 93--104.

\bibitem[{Zhang et~al.(2018)Zhang, Cisse, Dauphin, and
  Lopez-Paz}]{zhang2017mixup}
Hongyi Zhang, Moustapha Cisse, Yann~N Dauphin, and David Lopez-Paz. 2018.
\newblock \href {https://arxiv.org/abs/1710.09412} {mixup: Beyond empirical
  risk minimization}.
\newblock In \emph{International Conference on Learning Representations}.

\bibitem[{Zhang et~al.(2020)Zhang, Yu, and Zhang}]{zhang-etal-2020-seqmix}
Rongzhi Zhang, Yue Yu, and Chao Zhang. 2020.
\newblock \href {https://doi.org/10.18653/v1/2020.emnlp-main.691} {{S}eq{M}ix:
  Augmenting active sequence labeling via sequence mixup}.
\newblock In \emph{Proceedings of the 2020 Conference on Empirical Methods in
  Natural Language Processing (EMNLP)}, pages 8566--8579, Online. Association
  for Computational Linguistics.

\bibitem[{Zhu et~al.(2018)Zhu, Paschalidis, and Tahmasebi}]{zhu2018clinical}
Henghui Zhu, Ioannis~C Paschalidis, and Amir~M Tahmasebi. 2018.
\newblock \href {https://arxiv.org/abs/1810.10566} {Clinical concept extraction
  with contextual word embedding}.
\newblock In \emph{NIPS Machine Learning for Health Workshop}.

\end{thebibliography}
\bibliographystyle{acl_natbib}

\end{document}